\documentclass{article}

\PassOptionsToPackage{numbers, compress}{natbib}

\usepackage[final]{neurips_2022}

\usepackage[utf8]{inputenc} 
\usepackage[T1]{fontenc}    
\usepackage{hyperref}       
\usepackage{url}            
\usepackage{booktabs}       
\usepackage{amsfonts}       
\usepackage{nicefrac}       
\usepackage{microtype}      
\usepackage{xcolor}         
\usepackage{natbib}
\usepackage{multirow}
\usepackage{graphicx}
\usepackage{floatflt}
\usepackage{wrapfig}
\usepackage{tikz}
\usepackage{mathtools}
\usetikzlibrary{positioning,shapes,arrows}
\usepackage{caption}
\usepackage{subcaption}
\usepackage{float}

\newcommand{\key}{\textbf}

\title{A unified information-theoretic model of EEG signatures of human language processing}

\author{%
  Jiaxuan Li \qquad Richard Futrell\\
  Department of Language Science\\
  University of California Irvine\\
  Irvine, CA 92617 \\
  \texttt{\{jiaxuan.li,rfutrell\}@uci.edu} \\
}

\begin{document}

\maketitle

\begin{abstract}
    We advance an information-theoretic model of human language processing in the brain, in which incoming linguistic input is processed at two levels, in terms of a heuristic interpretation and in terms of error correction. We propose that these two kinds of information processing have distinct electroencephalographic signatures, corresponding to the well-documented N400 and P600 components of language-related event-related potentials (ERPs). Formally, we show that the information content (surprisal) of a word in context can be decomposed into two quantities: (A) \key{heuristic surprise}, which signals processing difficulty of word given its inferred context, and corresponds with the N400 signal; and (B) \key{discrepancy signal}, which reflects divergence between the true context and the inferred context, and corresponds to the P600 signal. Both of these quantities can be estimated using modern NLP techniques.  We validate our theory by successfully simulating ERP patterns elicited by a variety of linguistic manipulations in previously-reported experimental data from \citep{ryskin2021erp}. Our theory is in principle compatible with traditional cognitive theories assuming a `good-enough' heuristic interpretation stage, but with precise information-theoretic formulation.
\end{abstract}

\section{Introduction}
Human language comprehension is linked to (at least) two distinct and robust event-related potential (ERP) components detectable through electroencephalography---the N400 and P600. The N400 is a negative-going waveform that peaks at around 400 ms after the onset of linguistic signal, whereas the P600 is a positivity at around 600 ms. Since their discovery, a great deal of research has attempted to ascertain the functional interpretation of the N400 and P600 signals in order to shed light on the neural mechanisms of human language processing \citep[eg.][]{kutas_event-related_1980,hagoort_syntactic_1993,hoeks_seeing_2004,kim_independence_2005,van_herten_erp_2005,van2012prediction,kuperberg_neural_2007,kuperberg_separate_2016,van2012prediction}. 

Recent psycholinguistic theories have proposed a \key{heuristic interpretation} stage of language comprehension, where comprehenders form a plausible interpretation based on a subset of information in the input signal~\citep{kim_independence_2005,kuperberg_separate_2016,van_herten_erp_2005}. In such theories, the N400 reflects the degree of semantic mismatch in heuristic interpretation, and P600 indexes the effort of resolving conflicts between the heuristic interpretation and the veridical signal. 
This idea has been formalized in a noisy-channel framework, where comprehenders rationally infer a probabilistic distribution on the intended utterance given the received input while taking into account the fact that the input may contain errors (``noise'').
In support of this idea, prior work has established that there is a reduced N400 and a larger P600 when a plausible corrected sentence can be recovered from an original sentence containing a semantic error \citep{gibson_rational_2013,ryskin2021erp}.
However, none of the proposed theories can currently explain the full range of empirical ERP patterns (see~\citep{brouwer_getting_2012}), and existing models are not integrated with more general computational neuroscientific models.


We propose an information-theoretic computational-level model of the N400 and P600 ERP components in language processing, formalizing the noisy-channel intuition described above and integrating multiple strands of psycholinguistic research into a quantitative model that explains previously-reported results while making successful novel predictions about linguistic ERPs\footnote{We make all the codes and data are available at \url{https://anonymous.4open.science/r/NeurIPS-Workshop-Surprisal-Decomposition-ERP-CF9C/}.}. 

\section{Model}

Our model builds on Surprisal Theory, an empirically successful theory of behavioral signatures of language comprehension such as reading time \citep{hale2001probabilistic,levy2008expectation,frank2011insensitivity,smith2013effect,wilcox2020predictive}, which is in line with recent computational neuroscientific proposals to quantify cognitive effort information-theoretically \citep{ortega2013thermodynamics,zenon2019information,gershman2020origin,jakob2022rate,futrell2022information}. Surprisal Theory holds that the magnitude of processing effort for a word $x_t$ given a context of previous words $x_{<t}$ should be proportional to the information content or \key{surprisal} $S_t$ of the word given its context:
\begin{equation}
    S_t = -\ln p(x_t \mid x_{<t}).
\end{equation}
Our model maintains the idea that the total amount of processing effort is given by surprisal, but we partition the surprisal into two parts, corresponding to different forms of information processing and to the two distinct ERP signals.

\paragraph{Surprisal decomposition}

\begin{wrapfigure}{r}{0.5\textwidth}
    \centering
\begin{tikzpicture}[
  node distance=1cm ,
  mynode/.style={draw,ellipse,align=center}
]
\node[mynode] (T) {$T$};
\node[mynode,below of=T,left of=T] (Tp) {$W_{<t}$};
\node[mynode,below of=T,right of=T] (Tf) {$W_t$};
\node[mynode,shade, below of=Tp] (c) {$x_{<t}$};
\node[mynode,shade,below of=Tf] (w) {$x_t$};

\path (T) edge[-latex] (Tp);
\path (T) edge[-latex] (Tf);
\path (Tp) edge[-latex] (Tf);
\path (Tp) edge[-latex] (c);
\path (Tf) edge[-latex] (w);
\end{tikzpicture}
\caption{The comprehender's generative model. $T$ is the speaker's intended structure. At time $t$, the structure $T$ contains words $W_{<t}$ and (the past context) and $W_t$ (current word). The comprehender observes a noisy form of the past context $x_{<t}$ and the current word $x_t$.}
\label{fig:graphmodel}
\end{wrapfigure}
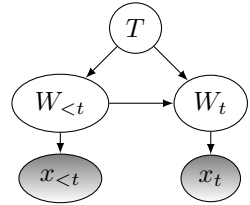

Consider a comprehender perceiving a sentence at time $t$, currently observing word $x_t$ in the context of (a memory trace of) previously-observed words $x_{<t}$. We formalize the idea of a `heuristic interpretation' in the generative model shown in Figure~\ref{fig:graphmodel}. Here the comprehender is trying to infer the value of a variable $T$ representing the speaker's intended structure, for example a complete parse tree. Crucially, the link between the intended structure $T$ and the input words $x$ is not deterministic: speakers may make errors in production, or environmental noise may disrupt the signal, and comprehenders should be able to correct for these factors. We formalize this idea by introducing random variables for \key{heuristic words} $W_{<t}$ and $W_t$, corresponding to the values of past words and the current word within the speaker's inteded structure $T$. The heuristic words give rise to the input words through a \textbf{noise model}, a distribution $p_N(x \mid W)$ representing all kinds of errors that might occur during language production and transmission.

We propose that, with each incoming word, the comprehender is updating her representations of the heuristic words $W$ and structure $T$. Within the generative model of Figure~\ref{fig:graphmodel}, the surprisal $S_t$ can be partitioned into two parts, corresponding to (A) the new information content of the heuristic words themselves, termed \key{heuristic surprise}, and (B) the update to beliefs about the heuristic words given the input words, termed \key{discrepancy signal}:
\begin{equation}
\label{eq:decomposition}
S_t =  \underbrace{\left\langle -\ln p(W_t \mid W_{<t}) \right\rangle}_{\text{heuristic surprise}, =A} + \underbrace{\left\langle \ln \frac{p(W_t \mid W_{<t})}{p(x_t \mid x_{<t})} \right\rangle}_{\text{discrepancy signal}, =B},
\end{equation}
where $\left\langle \cdot \right\rangle$ indicates an average with respect to the probability distribution $p(W_{\le t} \mid x_{\le t})$. The heuristic surprise is an upper bound on the information provided by the heuristic words about the structure $T$. We propose that the N400 magnitude is proportional to the heuristic surprise $A$ and the P600 magnitude is proportional to the discrepancy signal $B$ for distinct positive scalars $\alpha$ and $\beta$ in:
\begin{equation}
\label{eq:proportions}
\mathrm{N400} = \alpha A, \mathrm{P600} = \beta B,
\end{equation}



\paragraph{Noise model} 
The model quantities $A$ and $B$ are both averages with respect to the comprehender's probability distribution on heuristic words given input words, $p\left( W \mid x \right)$. This distribution can be written using Bayes' rule as 
\begin{equation}
    \label{eq:bayes}
    p(W \mid x) \propto p_N(x \mid W) p(W).
\end{equation}
To fully specify the model, therefore, requires us to specify (1) a noise model $p_N$ representing likely errors in production and/or transmission, and (2) a prior probability distribution $p(W)$, which reflects the probability that a speaker would want to produce a sequence of words $W$. 


\paragraph{Implementation}
To generate heuristic words for experimental stimuli, we follow Eq.~\ref{eq:bayes} applied independently to individual words. Here $p(W)$ for a single word is calculated using the Masked Language Model RoBERTa \citep{liu2019roberta}, and $p(x \mid W)$ for a single word is calculated as
\begin{equation}
\label{eq:levenshtein-falloff}
p(x \mid w) \propto e^{-\lambda d(x, w)},
\end{equation}
where $\lambda$ is a constant free parameter and $d(w, x)$ is the Levenstein edit distance between input word $x$ and heuristic word $w$. In order to generate candidate corrections, we replace target word in the input sentence with a special token \texttt{<mask>}, and use RoBERTa to generate the probability distribution to fill in the masked token. We selected top 100 predictions as our $W$. After that, we calculate the posterior probability by multiplying the RoBERTa probability with Eq.~\ref{eq:levenshtein-falloff}. We calculate the conditional probability of the current word $W_t$ given the context $W_{<t}$ with the autoregressive transformer GPT-2~\citep{radford2019language}. The conditional probability of veridical target $x_t$ given the veridical context $x_{<t}$ is also calculated with GPT-2.

\section{Empirical Validation}
\subsection{Dataset}
We validate our model using ERP data from~\citep{ryskin2021erp}, who report experiments designed to test how well noisy-channel error correction can explain linguistic ERP patterns.\footnote{The data is available at \url{https://osf.io/vcsfb/} without a license.} The experiments have four conditions (see Table~\ref{tab:stimuli}, one with a semantic violation (\emph{Sem}), one with a syntactic violation (\emph{Syn}), one semantic critical condition (\emph{SemCrit}) with a semantic violation which could be attributed to noise, and a control sentence without any error (\emph{Control}). The ERP effects in the three experimental conditions are all calculated in terms of differences to the ERP signal in the control condition. In the N400 time window, there is a significant N400 effect in \emph{Sem} and \emph{SemCrit} conditions, where the N400 effect in \emph{SemCrit} condition is reduced (see Fig.~\ref{fig:n400_human_400}. In the P600 time window, there is a significant P600 effect in \emph{Synt} and a smaller but significant P600 effect in \emph{SemCrit} condition (see Fig.~\ref{fig:p600_human_400}).

\begin{table}[!htb]
\centering
\begin{tabular}{lll}\toprule
\textbf{Condition} & \textbf{Sentence} & \textbf{Effect} \\\midrule
Semantic & The storyteller could turn any incident into an amusing \textit{hearse}. & N400 \\
Syntactic & The storyteller could turn any incident into an amusing \textit{anecdotes}. & P600 \\
SemCrit & The storyteller could turn any incident into an amusing \textit{antidote}. & N400, P600 \\
Control & The storyteller could turn any incident into an amusing \textit{anecdote}. & --- \\\bottomrule
\end{tabular}
\caption{List of conditions, sample sentences and ERP effects in dataset.}
\label{tab:stimuli}
\end{table}

\subsection{Simulation Results}
\begin{figure}[!htb]
     \centering
     \begin{subfigure}[b]{0.24\textwidth}
         \centering
         \includegraphics[width=\textwidth]{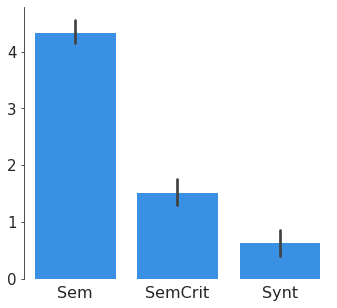}
         \caption{N400 (Human)}
         \label{fig:n400_human_400}
     \end{subfigure}
     \hfill
     \begin{subfigure}[b]{0.24\textwidth}
         \centering
         \includegraphics[width=\textwidth]{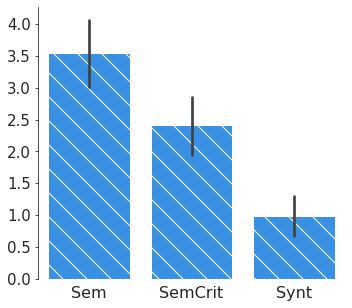}
         \caption{N400 (Simulated)}
         \label{fig:n400_roberta_400}
     \end{subfigure}
     \hfill
     \begin{subfigure}[b]{0.24\textwidth}
         \centering
         \includegraphics[width=\textwidth]{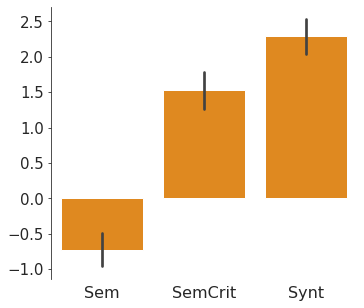}
         \caption{P600 (Human)}
         \label{fig:p600_human_400}
     \end{subfigure}
     \hfill
    \begin{subfigure}[b]{0.24\textwidth}
         \centering
         \includegraphics[width=\textwidth]{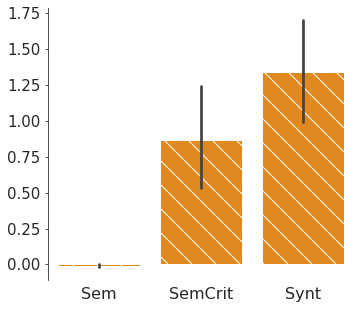}
         \caption{P600 (Simulated)}
         \label{fig:p600_roberta_400}
     \end{subfigure}
        \caption{N400 and P600 effect size from human experiment and from model prediction ($\lambda = 400$). }
        \label{fig:results}
\end{figure}

Fig.~\ref{fig:results} shows the simulated and empirical N400 and P600 effect sizes across conditions in the dataset, with $\lambda=400$ (see Appendix~\ref{sec:tuning} for results of a hyperparameter search on $\lambda$). The simulated effect sizes from the RoBERTa-based model implementation are similar to the effect sizes in real human ERP experiments. One potential issue in the use of language models for word probabilities is that, although large language models are sensitive to syntactic violations \citep{futrell2019syntactic,gauthier2020syntaxgym}, the surprisal penalty associated with syntactic violations is smaller than for semantic violations (see Appendix~\ref{sec:lm-surprisal}). This means that, in order to get corrections in the \emph{Synt} condition, it is necessary to set $\lambda$ to a very high value, 400. Model results with a smaller value of $\lambda=300$ are shown in Appendix~\ref{sec:tuning}; these show a better fit to the \emph{SemCrit} condition for the N400.

We statistically confirmed the relationship between the empirical ERP amplitudes (N400 and P600) and our information-theoretic measures (heuristic surprise $A$ and discrepancy signal $B$) in maximal linear mixed-effects models including by-subject and by-item intercepts and slopes \citep{barr2013random}. We find a significant main effect of heuristic surprise on N400 amplitude ($t = -6.57, p < .001$), and a significant main effect of discrepancy signal on P600 amplitude ($t = 3.64, p < .001$). In comparison, we find no significant effect of true surprisal on P600 ($t = 0.20, p = 0.84$), suggesting that our proposed decomposition of surprisal provides a better fit of the overall ERP components than true surprisal alone \citep{michaelov2021cloze,michaelov2020well}.

\begin{table}[H]
\centering
\caption{The effects of true surprisal $S_t$, heuristic surprise $A$, and discrepancy signal $B$ on ERP amplitudes in the N400 and P600 time range in the experiment from \citep{ryskin2021erp}. Numbers are $\beta$ values ($t$-values). \emph{p} $<$ 0.05*, \emph{p} $<$ 0.01**, \emph{p} $<$ 0.001***. A negative effect in the N400 range indicates the standard N400 effect; a positive effect in the P600 range indicates the standard P600 effect.}
\begin{tabular}{ccc@{\hspace{0.05cm}}cc}
\toprule
\multicolumn{2}{c}{N400}& & \multicolumn{2}{c}{P600} \\ \cmidrule{1-2} \cmidrule{4-5}
heuristic surprise & true surprisal  && discrepancy signal & true surprisal \\\midrule
-0.66 (-6.57***) & -0.59 (-5.76***) && 0.75 (3.64***) & 0.02 (0.20) \\\bottomrule
\end{tabular}
\label{tab:exp3_roberta}
\end{table}


\section{Conclusion}
We presented a neuro-computational model of N400 and P600 ERP components in language processing. We modeled the ERP components based on a generalized theory of surprisal. We argue that surprisal of word can be decomposed into two parts--- a heuristic surprise and a discrepancy signal, which correspond to N400 and P600 respectively. The two measures have a clear cognitive interpretation. The heuristic surprise signals the processing difficulty at the target word position given heuristic context, and the discrepancy signal represents the efforts of updating the discourse from inferred to true structure. We approximate the distribution on heuristic interpretations via a noisy-channel process, and implement it with large-scale language models. The theory is validated with experimental results. 

Our model provides an information-theoretic quantitative theory of language-related neural signals. By linking ERP components to Surprisal Theory, our model creates a precise formal link between theories of ERPs and other behavioral measures of language processing. The work calls for co-registration of brain and behavioral experiments to better understand the underlying cognitive process. 

Our model highlights the role of probabilistic inference in language processing, and provides a computational implementation of it. The noisy-channel model for heuristic interpretations abstracts away how different linguistic cues are weighted and combined by evaluating the heuristic interpretation based on a balance between prior belief and its divergence with new evidence. Furthermore, we leverage recent computational models from the field of natural language processing to implement our theory, which allows us to take into account the statistical variations in real experimental inputs. While we provide an implementation using pre-trained language models and edit distance, we want to acknowledge that this model is on a computational level, and further work could be done on the precise algorithmic nature of the heuristic interpretation generation process.


\bibliography{references}
\bibliographystyle{plain}

\section*{Checklist}


\begin{enumerate}

\item For all authors...
\begin{enumerate}
  \item Do the main claims made in the abstract and introduction accurately reflect the paper's contributions and scope?
    \answerYes{We propose and test an information-theoretic model of N400 and P600 in language processing.}
  \item Did you describe the limitations of your work?
    \answerYes{See the discussion of future work in the section Conclusion.}
  \item Did you discuss any potential negative societal impacts of your work?
    \answerNA{This work is concerned with cognitive process of language comprehension and adopts a computational approach. We cannot foresee any potential negative societal impacts}
  \item Have you read the ethics review guidelines and ensured that your paper conforms to them?
    \answerYes{}
\end{enumerate}

\item If you are including theoretical results...
\begin{enumerate}
  \item Did you state the full set of assumptions of all theoretical results?
    \answerNA{}
        \item Did you include complete proofs of all theoretical results?
    \answerNA{}
\end{enumerate}

\item If you ran experiments...
\begin{enumerate}
  \item Did you include the code, data, and instructions needed to reproduce the main experimental results (either in the supplemental material or as a URL)?
    \answerYes{see Appendix A.1. Data Availability.}
  \item Did you specify all the training details (e.g., data splits, hyperparameters, how they were chosen)?
    \answerYes{The section 2 Model and Appendix A2 provide model details.}
        \item Did you report error bars (e.g., with respect to the random seed after running experiments multiple times)?
    \answerYes{The fig 2. includes by-item error bars of empirical results and model predictions.}
        \item Did you include the total amount of compute and the type of resources used (e.g., type of GPUs, internal cluster, or cloud provider)?
    \answerNo{The computing can be done on a modern personal computer or google Colab with less than three hours.}
\end{enumerate}

\item If you are using existing assets (e.g., code, data, models) or curating/releasing new assets...
\begin{enumerate}
  \item If your work uses existing assets, did you cite the creators?
    \answerYes{}
  \item Did you mention the license of the assets?
    \answerNo{The assets are not licensed.}
  \item Did you include any new assets either in the supplemental material or as a URL?
    \answerNo{}
  \item Did you discuss whether and how consent was obtained from people whose data you're using/curating?
    \answerYes{The data are publicly available on \url{https://osf.io/vcsfb/}}
  \item Did you discuss whether the data you are using/curating contains personally identifiable information or offensive content?
    \answerNA{The data are either created by researchers or reported on publicly available journals with proper ethics reviews }
\end{enumerate}

\item If you used crowdsourcing or conducted research with human subjects...
\begin{enumerate}
  \item Did you include the full text of instructions given to participants and screenshots, if applicable?
    \answerNA{}
  \item Did you describe any potential participant risks, with links to Institutional Review Board (IRB) approvals, if applicable?
    \answerNA{}
  \item Did you include the estimated hourly wage paid to participants and the total amount spent on participant compensation?
    \answerNA{}
\end{enumerate}

\end{enumerate}


\appendix
\section{Appendix}
\subsection{Comparison between human- and LM-generated word probabilities.}\label{sec:lm-surprisal}
Table~\ref{tab:true_surpsial} shows the cloze probability obtained from human sentence completion task and from GPT-2 generations. All three experimental conditions (\emph{Sem}, \emph{SemCrit} and \emph{Synt}) have close to 0 human cloze probabilities, however, GPT-2 assigns a lower surprisal to the \emph{Synt} condition than the other two semantically anomalous conditions. The discrepancy between human and language model probabilities indicates that language models might have under-estimated the surprisal of syntactically anomalous sentences.
\begin{table}[htb!]
    \centering
    \caption{Averaged cloze probability and total surprisal of stimuli across conditions.}
    \begin{tabular}{ccccc}
    \toprule
         & \emph{Control} & \emph{Sem} & \emph{SemCrit} & \emph{Synt} \\\midrule
        Human cloze probability & 0.40 & 0 & 0 & 0\\\midrule
       GPT-2 generated surprisal  & 5.2 & 8.7 & 8.5 & 7.5\\\bottomrule
    \end{tabular}
    \label{tab:true_surpsial}
\end{table}

\subsection{Hyper-parameter Tuning}\label{sec:tuning}
We explored the effect of $\lambda$ with a grid search from 100 to 500, with a step size of 100, and with two marginal conditions ($\lambda = 0$ and $\lambda = 1000$). The simulated N400 and P600 across our selection of $\lambda$, together with true surprisal of the stimuli, are summarized in Table~\ref{tab:lambda_tuning}. When $\lambda = 0$, the heuristic surprise is the surprisal the most predictable words given the previous, regardless of the true target received. As $\lambda$ increases, it becomes more different to do error correction, with an increased heuristic surprise and a decreased error correction. Importantly, sentences in different conditions have different sensitivity to $\lambda$. Sentences in \emph{Synt} and \emph{SemCrit} have an easy fix, and therefore are more likely to be corrected given an increasingly large $\lambda$. After visual inspection, we chose $\lambda = 400$.

\begin{table}[H]
    \centering
    \caption{Averaged heuristic surprise and discrepancy signal under different $\lambda$ values.}
    \begin{tabular}{cccccc@{\hspace{0.05cm}}cccc}
    \toprule
       & \multicolumn{4}{c}{Heuristic Surprise ($\rightarrow$ N400)} && \multicolumn{4}{c}{Discrepancy Signal ($\rightarrow$ P600)}\\\cmidrule{2-5}\cmidrule{7-10}
       & \emph{Control} & \emph{Sem} & \emph{SemCrit} & \emph{Synt} && \emph{Control} & \emph{Sem} & \emph{SemCrit} & \emph{Synt}\\\midrule
    $\lambda = 0$  & 4.4 & 4.6 & 4.5 & 4.4 && 0.8 & 4.1 & 4.0 & 3.1 \\
    $\lambda = 100$  & 5.0 & 6.7 & 5.3 & 5.4 && 0.2 & 2.0 & 3.2 & 2.2 \\
    $\lambda = 200$  & 5.2 & 8.5 & 6.2 & 5.6 && 0.02 & 0.2 & 2.3 & 1.9 \\
    $\lambda = 300$  & 5.2 & 8.7 & 6.8 & 6.1 && 0.04 & 0.05 & 1.7 & 1.4 \\
    $\lambda = 400$  & 5.2 & 8.7 & 7.6 & 6.2 && 0.008 & 0 & 0.9 & 1.3 \\
    $\lambda = 500$  & 5.2 & 8.7 & 7.6 & 6.2 && 0.007 & 0 & 0.9 & 1.3 \\
    $\lambda = 1000$  & 5.2 & 8.7 & 8.5 & 7.5 && 0 & 0 & 0 & 0 \\\bottomrule
    \end{tabular}
    \label{tab:lambda_tuning}
\end{table}

\begin{figure}[ht!]
     \centering
     \begin{subfigure}[b]{0.24\textwidth}
         \centering
         \includegraphics[width=\textwidth]{exp3_human_n400_errorbar.png}
         \caption{N400 (Human)}
         \label{fig:n400_human_300}
     \end{subfigure}
     \hfill
     \begin{subfigure}[b]{0.24\textwidth}
         \centering
         \includegraphics[width=\textwidth]{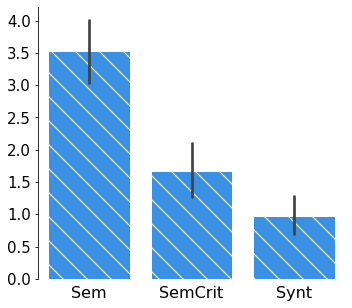}
         \caption{N400 (Simulated)}
         \label{fig:n400_roberta_300}
     \end{subfigure}
     \hfill
     \begin{subfigure}[b]{0.24\textwidth}
         \centering
         \includegraphics[width=\textwidth]{exp3_human_p600_errorbar.png}
         \caption{P600 (Human)}
         \label{fig:p600_human_300}
     \end{subfigure}
     \hfill
    \begin{subfigure}[b]{0.24\textwidth}
         \centering
         \includegraphics[width=\textwidth]{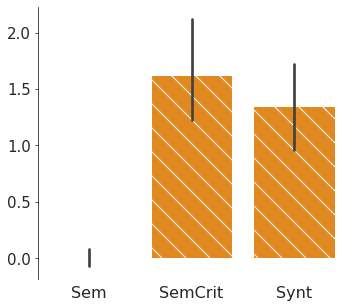}
         \caption{P600 (Simulated)}
         \label{fig:p600_roberta_300}
     \end{subfigure}
        \caption{N400 and P600 effect size from human experiment and from model prediction ($\lambda = 300$). }
        \label{fig:results_300}
\end{figure}

Fig.~\ref{fig:results_300} show a comparison between true ERP effect size in human experiment and simulated effect size when $\lambda = 300$. The model prediction aligns well with the real N400 effect size, but it underestimate the size of P600 effect. This is because GPT-2 underestimates the total surprisal of word when it has a syntactic violation.

\end{document}